%% file: paper.tex
\documentclass{bmvc2k}

\usepackage{indentfirst}
\usepackage{booktabs}
\usepackage{enumitem}
\usepackage{bbding}
\usepackage{array}
\newcolumntype{H}{>{\setbox0=\hbox\bgroup}c<{\egroup}@{}}

\title{Probabilistic Reconstruction Networks for 3D Shape Inference from a Single Image}
\addauthor{Roman Klokov}{roman.klokov@inria.fr}{1}
\addauthor{Jakob Verbeek}{jakob.verbeek@inria.fr}{1}
\addauthor{Edmond Boyer}{edmond.boyer@inria.fr}{1}

\addinstitution{
 Univ.\ Grenoble Alpes, Inria,\\
 CNRS, Grenoble INP*, LJK\\
 38000 Grenoble, France\\
*Institute of Engineering
}

\runninghead{Klokov, Verbeek, Boyer}{Probabilistic Reconstruction Networks}

\def\eg{\emph{e.g}\bmvaOneDot}
\def\ie{\emph{i.e}\bmvaOneDot}
\def\vs{\emph{vs}\bmvaOneDot}
\def\etc{\emph{etc}\bmvaOneDot}

\def\etal{\emph{et al}\bmvaOneDot}
\def\wrt{\emph{w.r.t}\bmvaOneDot}

\newcommand{\fig}[1]{Figure~\ref{fig:#1}}
\newcommand{\sect}[1]{Section~\ref{sect:#1}}

\newcommand{\tab}[1]{Table~\ref{tab:#1}}
\newcommand{\eq}[1]{Eq.~(\ref{eq:#1})}

\def\ex#1#2{\textrm{I\!E}_{#1}\!\left[#2\right]}     
\def\mypar#1{\vspace{1mm}{\noindent\bf #1.}\hspace{1mm}}

\def\figvspaceOne{\vspace{-5mm}}
\def\figvspaceTwo{\vspace{-5mm}}

\begin{document}
\maketitle
\input{abstract.tex}
\input{intro.tex}
\input{method.tex}
\input{related.tex}
\input{experiments.tex}
\input{conclusion.tex}
\input{appendices.tex}

\input{refs.tex}
\end{document}

%% file: abstract.tex
\begin{abstract}
We study end-to-end learning strategies for 3D shape inference from images, in particular from a single image. 
Several approaches in this direction have been investigated that explore different shape representations and suitable learning architectures. We focus instead on the underlying probabilistic mechanisms involved and contribute a more principled probabilistic inference-based reconstruction framework, which we coin Probabilistic Reconstruction Networks.
This framework expresses image conditioned 3D shape inference through a family of latent variable models, and naturally decouples the choice of shape representations from the inference itself. 
Moreover, it suggests different options for the image conditioning and allows training in two regimes, using either Monte Carlo or variational approximation of the marginal likelihood. 
Using our Probabilistic Reconstruction Networks we obtain single image 3D reconstruction results that set a new state of the art on the ShapeNet dataset in terms of the intersection over union and earth mover's distance evaluation metrics. 
Interestingly, we obtain these results using a basic voxel grid representation, improving over recent work based on finer point cloud or mesh based representations.
\end{abstract}

%% file: intro.tex
\section{Introduction}
\label{sect:intro}
The overwhelming success of convolutional neural networks on image data~\cite{lecun98ieee,krizhevsky12nips} instigated the exploration of CNNs for other problems, in particular in 3D visual computing. 3D CNNs for shapes represented with uniform voxel grids have  been investigated for  recognition~\cite{wu15cvpr, maturana15iros} and generative modelling tasks~\cite{brock16arxiv, wu16nips}. For 3D shape inference, initial works~\cite{girdhar16eccv, choy16eccv} successfully demonstrated the ability of 3D CNNs to produce coherent voxelized shapes given single images. This task has since gained a significant attention, as a result of its vast application field and despite its challenging ill-posed nature.

Further exploring CNNs in this context, recent works have investigated beyond straightforward adaptions of 2D CNNs to 3D voxel grids, notably to overcome the cubic complexity in time and memory associated with it. 
For instance, sparse representations of large voxel grid have been proposed to reduce complexity while allowing for finer shape details~\cite{tatarchenko17iccv,graham18cvpr}. 
Other more scalable shape representations suitable for recognition and generation tasks have been investigated, including rendered images \cite{su15iccv1}, geometry images \cite{sinha16eccv}, 2D depth maps~\cite{soltani17cvpr}, point clouds \cite{qi17cvpr,klokov17iccv, qi17nips}, and graphs \cite{monti17cvpr, verma18cvpr}. 
Importantly, these representations come with specific network architectures and loss functions suited to the corresponding data structures.

The variety of approaches complicates comparisons and identification of the sources of improved performance. 
In particular, most works do not decouple the problems inherent to the task, and mix new shape representations, along with the associated network architectures and loss functions, with different image conditioning schemes, different probabilistic formulations of the shape prediction task, and in some cases the use of additional training data. This leads to possibly difficult or even unfair comparisons, due to the inability to confidently determine the source of improvements in new models, and emphasizes the need for a more systematic approach.

\begin{figure}
  \begin{center}
  \scalebox{.7}{\includegraphics[width=\textwidth]{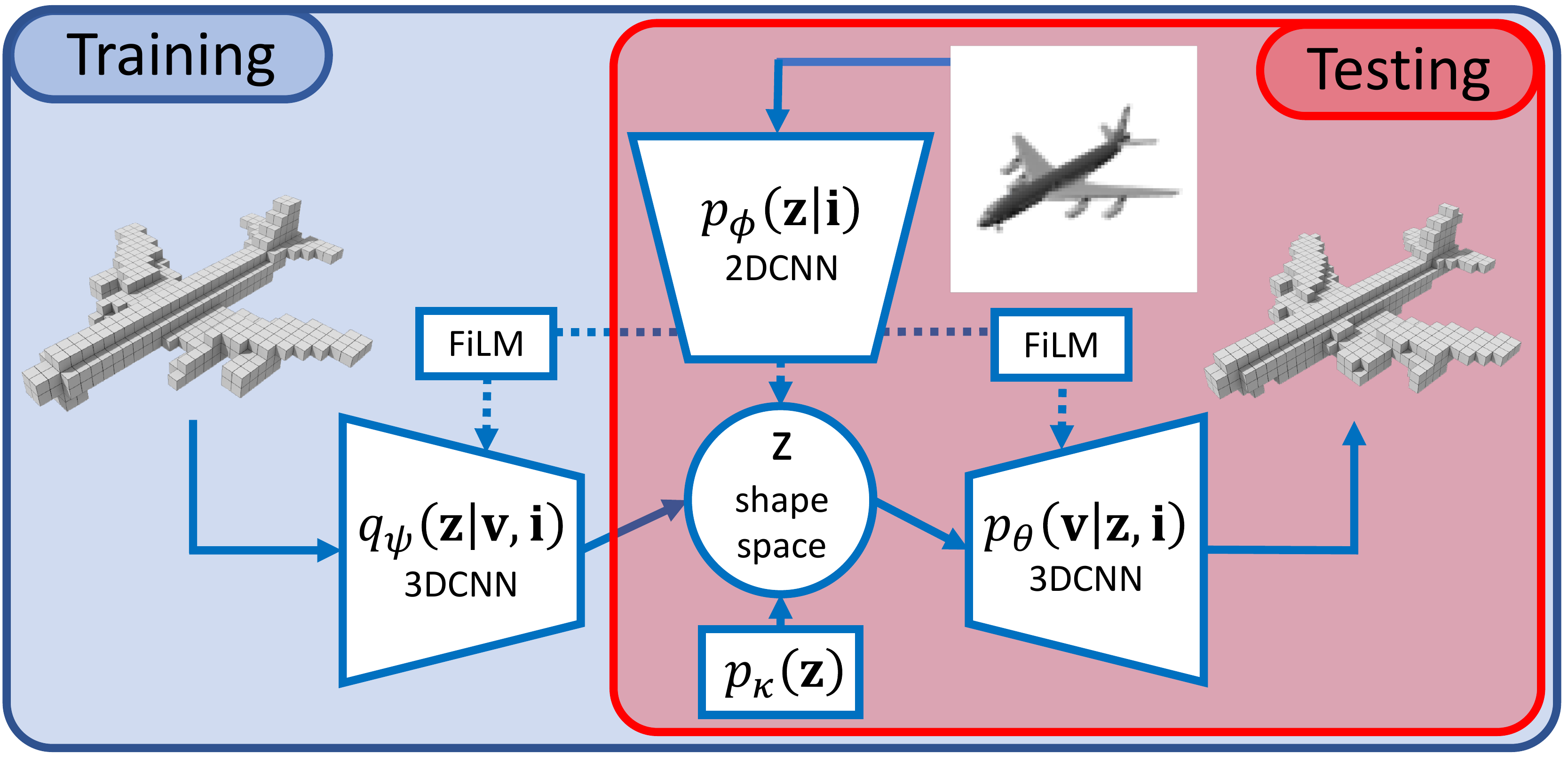}}
  \end{center}
  \figvspaceOne
  \caption{Probabilistic Reconstruction Networks for 3D shape inference from a single image. 
  Arrows show the computational flow through the model,  
  dotted arrows show optional image conditioning. 
  Conditioning between 2D and 3D tensors is achieved by means of FiLM~\cite{perez18aaai} layers.
  The inference network $q_\psi$ is only used during training for  variational inference.
  }
  \figvspaceTwo
  \label{fig:teaser}
\end{figure}

In this paper we look at the single image 3D shape inference through the prism of a family of generic probabilistic latent variable models, which we term {\it Probabilistic Reconstruction Networks} (PRN), see \fig{teaser}. 
The formalism encompassing these models naturally decouples different aspects of the problem including the shape representation, the image conditioning, and the usage of latent shape space for the shape prediction. 
It also allows to categorize previous models for this task by their structural properties.
Without loss of generality, we use voxel grids as shape representations and focus our attention on other aspects. 
We systematically analyze the impact of several design choices: 
(i) the dependency structure between the input image, the latent shape variable and the output variables; 
(ii) the effectiveness of training the model using Monte Carlo sampling or variational inference to approximate the log-likelihood;
(iii) the effectiveness of a deterministic version of the model that suppresses any uncertainty associated with the latent variable;
(iv) the effect of jointly learning the shape reconstruction model along with a generative shape model, which share their latent shape space. 

For our experiments we use the ShapeNet dataset for the single image 3D shape reconstruction. 
We obtain excellent single image 3D reconstruction results with our Probabilistic Reconstruction Networks, setting new state-of-the-art results in terms of the IoU and EMD performance metrics.
Interestingly, our results improve over recent works based on point-cloud and mesh-based shape representations. 

In summary, our contributions are:
\begin{itemize}[noitemsep,topsep=0pt]
  \item
  a generic latent variable model for the single image 3D shape reconstruction;
  \item
  exploration of modeling options in a systematic and comparable manner;
  \item
  new state-of-the-art single image reconstruction results on the ShapeNet dataset.
\end{itemize}
In the following we first introduce our generic latent variable model in \sect{method}, which is then used to review and categorize previous work on shape inference from a single image in \sect{related}. 
We then present our experimental results and comparisons in \sect{exp}.

%% file: method.tex
\section{Probabilistic framework for 3D shape reconstruction}
\label{sect:method}
Below we present our generic latent variable model in \sect{model}, and  detail the network architectures used for our experiments in \sect{archi}. 
Although proposed probabilistic model is agnostic to the underlying shape representation, we present it using voxel grid representation in accordance to our experimental setup.
\subsection{Latent variable model for single image 3D shape reconstruction}
\label{sect:model}
We consider a shape ${\bf v}$ as a uniform voxel occupancy grid of a predefined resolution. 
Our task is to predict the shape $\bf v$ given an input image ${\bf i}$, \ie to model $p\left({\bf v}|{\bf i}\right)$. 
While images and occupancy grids live in different spaces, both are representations of an underlying object that has a 3D shape and an appearance. Using this observation, we assume that an observed shape $\bf v$ has a latent parametrization $\bf z$ within a latent shape space of lower dimension, that captures shape variations.
We then define our latent variable model for single image 3D shape reconstruction as:
\begin{equation}
  \label{eq:pm_full}
  p\left({\bf v} | {\bf i}\right) = \int_{\bf z} p_{\theta}\left({\bf v} | {\bf z}, {\bf i}\right) p_{\phi}\left({\bf z} | {\bf i}\right) \text{d}{\bf z},
\end{equation}
where $\phi$ and $\theta$ are parameters of the model.
This model consists of two modules: an image informed latent variable prior $p_{\phi}\left({\bf z} | {\bf i}\right)$, and a decoder $p_{\theta}\left({\bf v} | {\bf z}, {\bf i}\right)$ that predicts the shape $\bf v$ given the image and the latent variables.  
Being a generic latent variable model, it allows us to decouple and study different aspects of the single image 3D shape reconstruction task.

\mypar{Image conditioning options} 
In \eq{pm_full} both latent variable prior and decoder are conditioned on the input image. 
If we drop the dependence on $\bf i$ from one module, we maintain dependence of the shape $\bf v$ on the image $\bf i$, and obtain two alternative models:
\begin{eqnarray}
  \label{eq:pm_latent}
  p\left({\bf v} | {\bf i}\right) & = &\int_{\bf z} p_{\theta}\left({\bf v} | {\bf z}\right) p_{\phi}\left({\bf z} | {\bf i}\right) \text{d}{\bf z},\\
  \label{eq:pm_cond}
  p\left({\bf v} | {\bf i}\right) & = &\int_{\bf z} p_{\theta}\left({\bf v} | {\bf z}, {\bf i}\right) p_{\phi}\left({\bf z}\right) \text{d}{\bf z}.
\end{eqnarray}
In the first case, we omit the image conditioning in the decoder $p_{\theta}\left({\bf v} | {\bf z}\right)$ and assume that the image conditioned prior $p_{\phi}\left({\bf z} | {\bf i}\right)$ is sufficient to obtain a valid reconstruction by the decoder. 
This dependency structure corresponds to an assumption of conditional independence of $\bf i$ and $\bf v$ given $\bf z$.
In the second case, we leave the image dependence in the decoder $p_{\theta}\left({\bf v} | {\bf z}, {\bf i}\right)$ but use an unconditional prior for the latent variable $p_{\phi}\left({\bf z}\right)$.
This corresponds to an assumption that $\bf i$ and  $\bf z$ are a-priori independent. 
If we drop the image conditioning in both components, the model becomes a generative latent variable shape model: 
$p\left({\bf v}\right) = \int p_{\theta}\left({\bf v} | {\bf z}\right) p_{\kappa}\left({\bf z}\right) \text{d}{\bf z}.$

\mypar{Latent variable sampling during training} 
Due to the non-linear dependencies in the integral in the models defined in equations (\ref{eq:pm_full})--(\ref{eq:pm_cond}), exact computation of the log-likelihood and its gradient is intractable. 
We consider two alternative approaches to overcome this difficulty. 
The first is to use a Monte Carlo approximation, as \eg in \cite{gordon19iclr},
\begin{eqnarray}
 \ln p({\bf v}|{\bf i})\approx \ln \frac{1}{M}\sum_{m=1}^M p_\theta({\bf v}|{\bf z}_m), \quad\quad\quad {\bf z}_m\sim p_\phi({\bf z}| {\bf i}),
\end{eqnarray}
where we make use of the re-parametrization trick~\cite{kingma14iclr,rezende14icml} to back-propagate the gradient of the log-likelihood \wrt $\phi$ through the sampling from $p_\phi({\bf z}| {\bf i})$. 

Alternatively, we can use the variational inference framework~\cite{kingma14iclr,rezende14icml} to obtain a training signal based on more informed samples from the latent variable. 
We introduce a variational approximate posterior $q_{\psi}\left({\bf z} | {\bf v}, {\bf i}\right)$, which is learned jointly with the prior and decoder by the maximization of the variational lower bound on the log-likelihood:
\begin{equation}
  \label{eq:elbo_cond}
  \mathcal{L}\left(\phi, \psi, \theta, {\bf v}, {\bf i}\right) = \ex{q_{\psi}\left({\bf z} | {\bf v}, {\bf i}\right)}{\log p_{\theta}\left({\bf v} | {\bf z}, {\bf i}\right)} - \mathcal{D}_\text{KL}\left(q_{\psi}\left({\bf z} | {\bf v}, {\bf i}\right) || p_{\phi}\left({\bf z} | {\bf i}\right)\right)\leq \ln p({\bf v}|{\bf i}).
\end{equation}
To evaluate this bound and its gradient, we sample from $q_{\psi}\left({\bf z} | {\bf v}, {\bf i}\right)$. 
Since the samples are conditioned on the shape, unlike the Monte Carlo approximation case, we expect this approach to be more sample efficient. 
On the one hand, image conditioning in the posterior may be omitted, since the posterior is already conditioned on the shape information. 
On the other hand, conditioning on the image may in principle further improve the accuracy of the approximate posterior, as it is based on more information.

\mypar{Deterministic shape model} 
We can obtain deterministic versions of the presented models by considering ${\bf z}$ as a function of $\phi$ and, if required, ${\bf i}$. 
Although this simplifies the models, it also discards an important property. 
Typically, $p({\bf v}|{\bf z})$ is factorized, with each voxel occupancy being modelled as an independent Bernoulli, \eg \cite{choy16eccv,girdhar16eccv}. 
With this factorization, any shape ambiguity given a single image cannot be modelled properly, since self-occluded parts of the shape lack structure in the prediction. 
A latent variable that conditions the factorized distribution can be used to induce dependencies among the voxel occupancies to reflect the structured ambiguity resulting from partially observed shapes.

In case of the variational training we also use a deterministic posterior, and substitute the KL-divergence in \eq{elbo_cond} with a suitable similarity measure. 
For example, the $L_2$-norm of the difference between the output of image encoder $p_\phi$ and the posterior $q_\psi$.

\mypar{Merging unconditional generation with reconstruction} 
The models presented above can be trained along with a generative shape model, that may be of interest on its own, or used to regularize the conditional model. 
To achieve this, we consider a variational bound on the log-likelihood of an unconditional generative model:
\begin{equation}
  \label{eq:elbo_uncond}
  \mathcal{L}\left(\kappa, \psi, \theta, {\bf v}\right) = \ex{q_{\psi}\left({\bf z} | {\bf v}\right)}{\log p_{\theta}\left({\bf v} | {\bf z}\right)} - \mathcal{D}_\text{KL}\left(q_{\psi}\left({\bf z} | {\bf v}\right) || p_{\kappa}\left({\bf z}\right)\right) \leq \ln p({\bf v}),
\end{equation}
where $p({\bf v})=\int p_\theta({\bf v}|{\bf z})p_\kappa({\bf z})  \text{d}{\bf z}$.
Looking at \eq{elbo_cond} and \eq{elbo_uncond}, we observe that if we omit image conditioning from the decoder and the posterior in \eq{elbo_cond}, and share their parameters in both conditional and unconditional models, we can obtain a unified training objective:
\begin{equation}
  \label{eq:elbo_comb_0}
  \begin{split}
    \mathcal{L}\left(\kappa, \phi, \psi, \theta, {\bf v}, {\bf i}\right) = \ex{q_{\psi}\left({\bf z} | {\bf v}\right)}{\log p_{\theta}\left({\bf v} | {\bf z}\right)} &- \frac{1}{2}  \mathcal{D}_\text{KL}\left(q_{\psi}\left({\bf z} | {\bf v}\right) || p_{\kappa}\left({\bf z}\right)\right) \\
    &- \frac{1}{2} \mathcal{D}_\text{KL}\left(q_{\psi}\left({\bf z} | {\bf v}\right) || p_{\phi}\left({\bf z} | {\bf i}\right)\right),
  \end{split}
\end{equation}
which is the average of the lower bounds on the marginal likelihood on the shape $\bf v$ with and without conditioning on the image.
The term corresponding to the unconditional likelihood can be viewed as a regularization that encourages $\bf z$ from the entire latent space to correspond to realistic shapes, instead of just the $\bf z$ in the support of the conditional distributions $p({\bf z}|{\bf i})$ on latent coordinates given an input image.
Note that in the generative model neither the decoder $p_{\theta}\left({\bf v} | {\bf z}\right)$ nor the encoder $q_{\psi}\left({\bf z} | {\bf v}\right)$ are conditioned on the image. 
These can thus be shared with the single image reconstruction model if the latter is not conditioned on the image in these components.

\vspace{-0.2cm}
\subsection{Network architectures}
\label{sect:archi}
Although our probabilistic model is not representation specific, we focus on the voxel grid shape representation, and leave the comparison to alternative representations for the future work. 
Thus, we implement the different conditional distributions in our model as 2D and 3D CNNs that output the parameters of distributions on the latent variable $\bf z$, or on the voxel occupancies. 
In particular, 
\begin{itemize}[noitemsep,topsep=0pt]
\item 
The unconditional Gaussian latent prior $p_{\kappa}\left({\bf z}\right)$ is characterized by $\kappa$ that consists of means and variances for all latent dimensions implemented as trainable parameters.
\item 
The image conditioned prior $p_{\phi}\left({\bf z} | {\bf i}\right)$ is a 2D CNN, consisting of six blocks of pairs of convolutions: a standard and a strided one, interleaved with batch normalization and point-wise non-linearities, followed by two fully-connected layers. 
It processes input images into the means and variances of a factored Gaussian on $\bf z$.
\item
The shape conditioned variational posterior $q_{\psi}\left({\bf z} | {\bf v}, {\bf i}\right)$  is a 3D CNN consisting of an initial 3D convolution and a series of four modified residual blocks~\cite{he16cvpr,he16eccv}, each using an additional 1$\times$1 convolution instead of identity and feature map concatenation instead of summation and each followed by 2$\times$2$\times$2 spatial average pooling. 
Final convolutional features are fed to two additional fully-connected layers. 
This encoder processes input shapes into the means and variances of a factored Gaussian on $\bf z$.
\item 
The 3D deconvolutional decoder $p_{\theta}\left({\bf v} | {\bf z}, {\bf i}\right)$ is mirrored from the approximate posterior $q_\psi$, with the pooling being substituted by the 2$\times$2$\times$2 upscaling by trilinear interpolations, producing the parameters of Bernoulli distributions on the voxel occupancies from a latent variable input.
\end{itemize}
Image conditioning in the two latter modules is inspired by the FiLM conditioning mechanism~\cite{perez18aaai}.
Intermediate 2D feature maps from the first five convolutional blocks of the image encoder $p_\phi({\bf z}|{\bf i})$ are averaged spatially, transformed by two additional fully-connected layers into weights and biases, that are used to scale the according five intermediate batch-normalized 3D feature maps in the shape encoder, the latent variable decoder, or both. 
Instead of affine transformation used in FiLM, we use non-negative scaling weights by predicting them in logarithmic scale, in our experiments this resulted in more stable training and slightly better results. 
See \fig{teaser} for a schematic overview of the model architecture.

To ensure fair comparison between the variations of the model we use identical architectures for every component concurring in different models, except for additional fully-connected layers associated with the different image conditioning options. 
When we include an unconditional generative shape model and optimize \eq{elbo_comb_0}, we share the decoder $p_\theta({\bf v}|{\bf z})$ and the variational posterior $q_\psi({\bf z}|{\bf v})$ between the conditional and unconditional models. Exact architectures, training procedures and their hyperparameters are available on the implementation code page.\footnote{\url{https://github.com/Regenerator/prns}}

%% file: related.tex
\section{Related work}
\label{sect:related}

In this section we review related work on single image 3D shape reconstruction, and relate it to our generic latent variable model presented in the previous section.

\begin{table*}
  \begin{center}
    \input{model_comparison.tex}
  \end{center}
  \caption{Overview of how related work fits into our probabilistic reconstruction framework.}
  \figvspaceTwo
  \label{tab:models}
\end{table*}

\mypar{Representations for 3D shape inference} 
The majority of works studying the inference-based single image 3D shape reconstruction introduce new shape representations and suitable neural network architectures. 
The seminal works by Choy \etal~\cite{choy16eccv} and Girdhar \etal~\cite{girdhar16eccv} used 3D CNNs to predict voxel occupancy grids. 
To reduce the computational complexity of the voxel grid representation Tatarchenko \etal~\cite{tatarchenko17iccv} proposed an architecture to  process octrees computed on top of the voxel grids. 
Richter and Roth~\cite{richter18cvpr} proposed to use a set of six depth maps to represent voxel grids and to combine a series of such sets in a nested manner to model non-trivial shapes. 
Su \etal~\cite{su17cvpr} combined 2D CNNs with fully-connected networks to output point clouds, which are learned by optimizing Chamfer distance or differentiable approximation of earth mover's distance. 
Wang \etal~\cite{wang18eccv} applied graph-convolutional networks \cite{bronstein17spm} to the mesh-based shape representation. 
Shin \etal~\cite{shin18cvpr} proposed to predict multiple depth maps and according silhouettes and fuse them into meshes by post-processing with Poisson reconstruction algorithm.

Although related in their overall goals, these approaches are difficult to compare 
since they use target shapes approximated to different degrees. 
Ideally, a fair comparison across shape representations should be performed while maintaining the same level of granularity across representations, and for different levels, since it is possible that some representations work better for rough shape reconstruction, while other are best for detailed reconstruction.

\mypar{Image-shape consistency and additional data} Another significant stream of works originates from the idea of ensuring consistency between input data and target shapes. 
Initial work by Yan \etal~\cite{yan16nips} introduced the consistency between 3D shapes and their silhouettes produced by different viewpoints in a form of a loss function. 
Similar ideas were investigated by Wiles and Zisserman~\cite{wiles17bmvc}.
Tulsiani \etal~\cite{tulsiani17cvpr} expanded this approach by the use of differentiable ray tracing in the loss function, ensuring correspondence of inferred voxelized shapes to foreground segmentation masks and depth images. 
Wu \etal~\cite{wu17nips} developed the idea even further and introduced a two-step reconstruction framework. 
The first part of the model is trained to infer 2.5D shape sketches (unions of segmentation, depth and normal maps) from images, while the second is separately trained to predict shapes from 2.5D sketches. 
Both components are then fine-tuned, using reprojection consistency.

Henderson and Ferrari~\cite{henderson18bmvc} proposed a probabilistic framework for image generation conditioned on a latent shape variable and an additional latent variable for the shape pose. This framework was used to train an underlying 3D mesh generator with the help of differentiable rendering of 3D meshes into images.
Differentiable point clouds~\cite{insafutdinov18nips} closed the consistency loop between inferred point clouds and input images, by rendering point clouds as images and minimizing loss between such renderings and input images. 

Similarly to the previous class of models, these methods also enrich available training data by considering different forms of additional data: camera information, 2.5D sketches, \etc. 
This, again, makes comparison problematic, since it is not always clear what the impact of the additional training data is.

\mypar{Relations to our framework} 
In addition to the work discussed above, given similarities between VAEs and GANs~\cite{hu18iclr}, our work is also related to adversarial approaches involving shape discriminators~\cite{wu16nips, smith17corl, wu18eccv}. 
In \tab{models} we organize related work in terms of how it fits into our generic probabilistic reconstruction framework, abstracting away from the   implementation of various components.

We see that most previous works use a dependency structure where the latent variable is inferred from the image, and the shape decoder only depends on the latent variable and not on the image.
Moreover, most works rely on deterministic models, except for GAN-based approaches of \cite{wu16nips, smith17corl}, the point cloud based approaches of \cite{su17cvpr, mandikal18bmvc}, and the mesh based method of \cite{henderson18bmvc}.
Finally, only TL-Networks \cite{girdhar16eccv} and 3D-LMNet \cite{mandikal18bmvc} make use of the variational inference for shape modelling.

%% file: model_comparison.tex
\scalebox{.9}{
\resizebox{\textwidth}{!}{
\begin{tabular}{lccc|c}
  \toprule
  Dependencies & Sampling & Deterministic & Discriminator & References\\
  \midrule
  $p\left({\bf s}|{\bf z}\right)p\left({\bf z}|{\bf i}\right)$  & $p\left({\bf z}|{\bf i}\right)$ & \footnotesize{\Checkmark} &                                        & \cite{choy16eccv, yan16nips, tulsiani17cvpr, wiles17bmvc, tatarchenko17iccv, wu17nips, richter18cvpr, shin18cvpr,  insafutdinov18nips}\\
  $p\left({\bf s}|{\bf z}\right)p\left({\bf z}|{\bf i}\right)$  & $p\left({\bf z}|{\bf i}\right)$ &                           &                                        & \cite{henderson18bmvc}\\
  $p\left({\bf s}|{\bf z}, {\bf i}\right)p\left({\bf z}\right)$ & $p\left({\bf z}\right)$         & \footnotesize{\Checkmark} &                                        & \cite{wang18eccv}\\
  $p\left({\bf s}|{\bf z}, {\bf i}\right)p\left({\bf z}\right)$ & $p\left({\bf z}\right)$         &                           &                                        & \cite{su17cvpr}\\
  \hline
  $p\left({\bf s}|{\bf z}\right)p\left({\bf z}|{\bf i}\right)$  & $p\left({\bf z}|{\bf i}\right)$ &                           & \footnotesize{\Checkmark} & \cite{wu16nips, smith17corl}\\
  $p\left({\bf s}|{\bf z}\right)p\left({\bf z}|{\bf i}\right)$  & $p\left({\bf z}|{\bf i}\right)$ & \footnotesize{\Checkmark} & \footnotesize{\Checkmark} & \cite{wu18eccv}\\
  \hline
  $p\left({\bf s}|{\bf z}\right)p\left({\bf z}|{\bf i}\right)$  & $q\left({\bf z}|{\bf s}\right)$ & \footnotesize{\Checkmark} &                                        & \cite{girdhar16eccv}\\
  $p\left({\bf s}|{\bf z}\right)p\left({\bf z}|{\bf i}\right)$  & $q\left({\bf z}|{\bf s}\right)$ &                           &                                        & \cite{mandikal18bmvc}\\
  \bottomrule
\end{tabular}
}
}

%% file: experiments.tex
\vspace{-3mm}
\section{Experiments}
\label{sect:exp}

In this section we present the experimental setup, our quantitative and qualitative evaluation results, as well as our analysis of these results.

\subsection{Dataset, evaluation metrics, and experimental details}
\mypar{Dataset} We evaluate PRNs on a subset of the ShapeNet dataset \cite{chang15arxiv1} introduced by Choy \etal~\cite{choy16eccv}. 
It contains about 44k 3D shapes from 13 major categories of ShapeNet dataset represented as voxel grids of resolution $32^3$, as well as renderings from 24 different randomized viewpoints as $137^2$ images. 
Following Choy \etal, we use 80\% of the shapes from each category for training and remaining shapes for testing.

\begin{table*}
  \begin{center}
    \input{results_internal.tex}
  \end{center}
  \caption{Evaluation results for variations of PRN. 
  Monte Carlo training uses samples from the unconditional or image-informed prior, while variational training relies on samples from the shape-conditioned approximate posterior. We report IoU under two occupancy probability thresholds $\tau$.}
  \figvspaceTwo
  \label{tab:res_int}
\end{table*}

\mypar{Evaluation metrics} We evaluate using the standard intersection-over-union (IoU) metric~\cite{choy16eccv}, which averages the per-category IoU metric between the inferred shape and the ground-truth voxel representation. 
In addition, to allow comparison to recent work based on point-cloud and mesh based representations, we also report the Chamfer distance (CD) and earth mover's distance (EMD), computed using the code of~\cite{sun18cvpr}, where we removed the square root from the distance computations in CD to make it comparable to the related work.
In particular, each ground truth and predicted voxel grid is mapped to a point cloud by sampling the surface induced using the marching cubes algorithm~\cite{lewiner03ggg}. 
We then compute the CD and EMD on the resulting point clouds.

\mypar{Training and evaluating} 
When using either Monte Carlo approximation or a variational objective function, we always use a single sample to compute the gradients during training. 
A unified training protocol is used for all the models: all components are trained simultaneously (contrary to \cite{girdhar16eccv, wu18eccv, mandikal18bmvc}), with the AMSGrad~\cite{reddi18iclr} optimizer with decoupled weight decay regularization~\cite{loshchilov19iclr} with step-like scheduling for learning rate and weight decay parameter, and restarts of gradient moments accumulation at the beginning of each step.

During testing, we use a deterministic approach. In particular, we take the means of the conditional distributions rather than samples from them. 
We found this to significantly improve the reconstruction quality, compared to sampling.

\subsection{Experimental results}
\mypar{Evaluation of PRN variants}
To explore the various possibilities of our general latent variable model, we consider three options to condition on the image: 
(i) using the latent space to carry all image information: $p({\bf v}|{\bf z})p({\bf z}|{\bf i})$, 
(ii) using additional conditioning of the decoder on the image: $p({\bf v}|{\bf z,i})p({\bf z}|{\bf i})$, and 
(iii) using an uninformed prior on the latent variable: $p({\bf v}|{\bf z,i})p({\bf z})$. 
To train the models we either approximate the integral in \eq{pm_full} directly with Monte Carlo samples from the prior on $\bf z$, or with a variational lower bound and samples from the variational posterior.
We also test a deterministic model, where the distribution on the latent variable is replaced by a deterministic function. 
Finally, we consider the option to train the model jointly with an an  unconditional generative model.
In \tab{res_int} we present the results, using two thresholds $\tau$ on the voxel occupancy probability: the neutral 0.5, and following \cite{choy16eccv} the looser 0.4 which overall leads to improved IoU scores.

In case of the Monte Carlo approximation (top three rows), additional image conditioning in the decoder improves the results.
Conditioning both the latent variable prior and the decoder on the image achieves best results, suggesting that these different pathways to use the image are complementary.

The use of variational training consistently improves the results over the Monte Carlo approximation. 
In this case, the additional image conditioning of the decoder or the variational posterior, see line five and six, is not effective and even somewhat reduces the performance. 
This is contrary to the results obtained with Monte Carlo sampling; in the latter case the sampling inefficiency is probably partially compensated by the additional conditioning pathway.
Variational inference leads to more accurate samples, which obviates the need for the additional image conditioning (at least for the chosen mechanism of the additional image conditioning).

We also consider a deterministic variant of our best performing model, which resembles the TL-network of~\cite{girdhar16eccv}.
The results show that probabilistic handling of the latent variable reduces overfitting in the model and leads to IoU of 2.5 points higher.
Finally, we also tested the training with a joint generative shape model, which TL-Networks used as pre-training. 
Although we did not observe a significant effect due to the joint training with a generative shape model, it does offer additional functionality by being able to sample shapes, or compute their likelihoods under the model.

\begin{table*}
  \begin{center}
    \input{results_external.tex}
  \end{center}
  \caption{Comparison of PRN to the state-of-the-art. 
  All results are taken from the original papers, except for $\dagger$, which were provided in~\cite{su17cvpr}. 
  Pixel2Mesh additionally uses camera information and surface normals during training.
  }
  \figvspaceTwo
  \label{tab:res_ext}
\end{table*}

\mypar{Comparison to the state-of-the-art} 
In \tab{res_ext} we compare to earlier state-of-the-art approaches.  
All methods use the same input images, but use slightly different image preprocessing: 3D-R2N2 uses random cropping, 3D-LMNet central cropping, AtlasNet crops and resizes, while PSGN and Pixel2Mesh resize the image. 
As OGN, we use original images, but also add a grey-scale version of each image as a fourth input channel.

In \tab{res_ext} we report Chamfer distance computed for 1024 predicted and 1024 ground truth points as it was done in most of the related work. Although they all used different numbers of predicted and ground truth points for training, The authors of \cite{groueix18cvpr, wang18eccv, mandikal18bmvc} explicitly state this protocol for evaluation, while there is no information about it in the text of \cite{su17cvpr}. Judging from the code of \cite{su17cvpr}, we assume that, in their case, the metric was obtained for 1024 predicted and 16384 ground truth points. For the reference, we recomputed CD under the same protocol and obtained a better value: 3.90. This shows that the metric is affected by a negative bias, which decreases with the increasing number of evaluated points, and underlines the need for unified evaluation protocol.

Our PRN obtains excellent results, and significantly improves over previous state-of-the-art results in terms of IoU and EMD, including methods based on point cloud and mesh representations. 
Point-based approaches use loss functions based on the Chamfer distance, and so naturally perform well in terms of this metric, but this does not per se transfer to better performance in the other metrics. 
In our case, we do not explicitly optimize for either of these metrics, relying on the binary cross-entropy for the voxel occupancies instead, and yet obtain best results in two of the three metrics (with only one competitor being better in terms of the third metric).

\begin{figure}
  \begin{center}
  \scalebox{0.9}{\includegraphics[width=\textwidth]{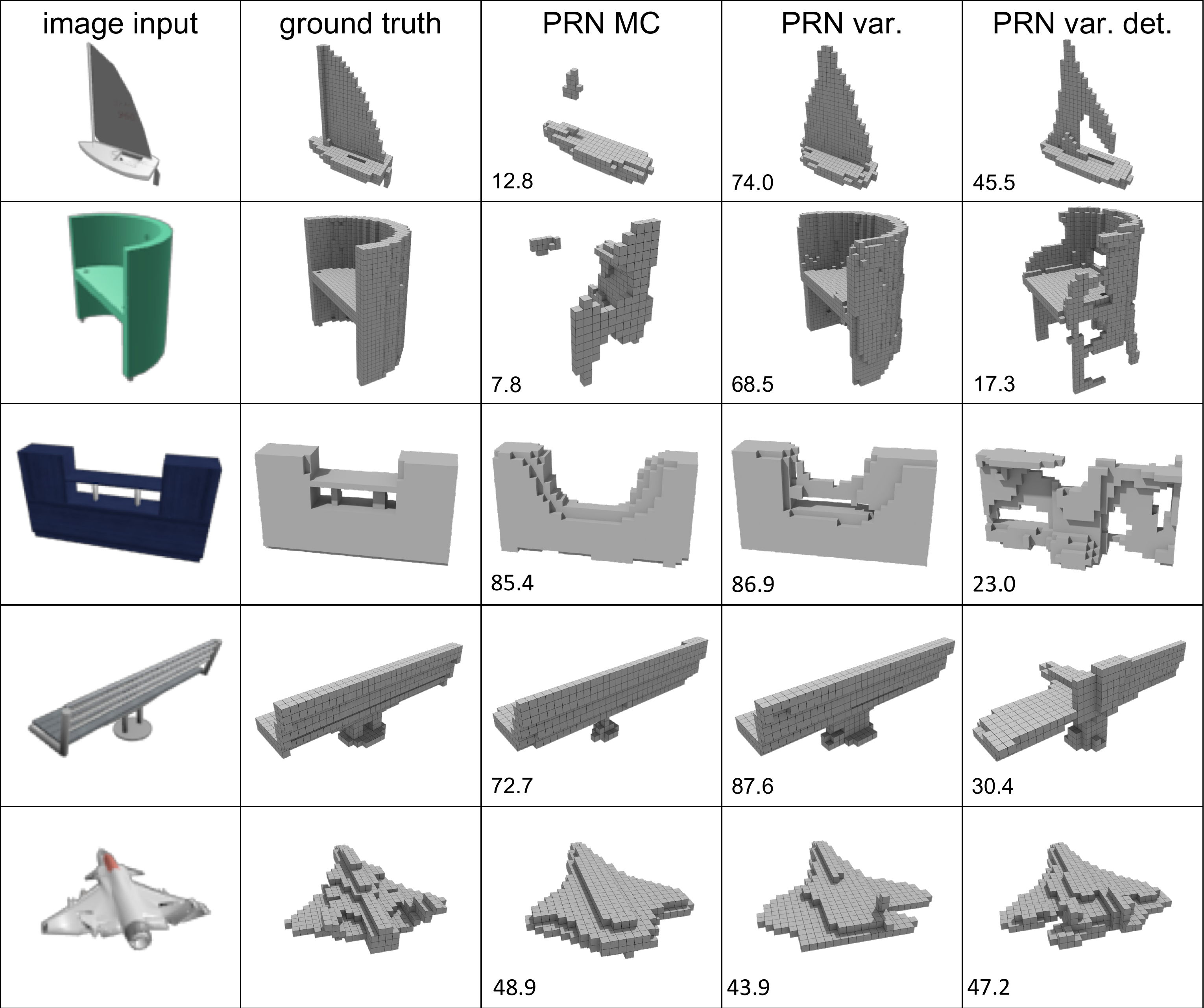}}
  \end{center}
  \figvspaceOne
  \caption{Qualitative reconstruction results for three variants of PRNs.}
  \figvspaceTwo
  \label{fig:examples}
\end{figure}

\mypar{Qualitative reconstruction results} 
In \fig{examples} we provide a selection of qualitative reconstruction results. 
We show results for the models in rows one, four and seven in \tab{res_int}, \ie with the $p({\bf v}|{\bf z})p({\bf z}|{\bf i})$ dependency structure, using Monte Carlo (PRN MC) and variational (PRN var.) training, and the deterministic version of the latter (PRN var.\ det.).
We show four examples where the variationally trained model is the best, and one case where it is the worst. 
Overall, variationally trained model output fewer failed reconstructions, as well as more detailed reconstructions, compared to more failures and over simplified reconstructions from the model trained with Monte Carlo.
For reference, the average IoU score of the variationally trained model is 66.2 (median 69.9), which corresponds to a fairly accurate reconstruction level, in particular given the challenging nature of the task.

%% file: results_internal.tex
\scalebox{.9}{
\resizebox{\textwidth}{!}{
\begin{tabular}{lccccc}
  \toprule
  Dependencies & Sampling & Deterministic &  Shape model & IoU$\uparrow (0.5)$ & IoU$\uparrow (0.4)$\\
  \midrule
  $p\left({\bf v}|{\bf z}\right)p\left({\bf z}|{\bf i}\right)$          & $p\left({\bf z}|{\bf i}\right)$ &  &  & 63.7       & 65.0\\
  $p\left({\bf v}|{\bf z}, {\bf i}\right)p\left({\bf z}|{\bf i}\right)$ & $p\left({\bf z}|{\bf i}\right)$ &  &  & 64.6       & 65.6\\
  $p\left({\bf v}|{\bf z}, {\bf i}\right)p\left({\bf z}\right)$         & $p\left({\bf z}\right)$         &  &  & 64.0       & 65.0\\
  \hline
  $p\left({\bf v}|{\bf z}\right)p\left({\bf z}|{\bf i}\right)$          & $q\left({\bf z}|{\bf v}\right)$ &  &  & {\bf 65.9} & {\bf 66.2}\\
  $p\left({\bf v}|{\bf z}, {\bf i}\right)p\left({\bf z}|{\bf i}\right)$ & $q\left({\bf z}|{\bf v}\right)$ &  &  & 64.8       & 65.3\\
  $p\left({\bf v}|{\bf z}\right)p\left({\bf z}|{\bf i}\right)$          & $q\left({\bf z}|{\bf v}, {\bf i}\right)$ &            & & 65.4 & 65.8\\
  \hline
  $p\left({\bf v}|{\bf z}\right)p\left({\bf z}|{\bf i}\right)$          & $q\left({\bf z}|{\bf v}\right)$          & \Checkmark & & 63.4 & 63.7\\
  $p\left({\bf v}|{\bf z}\right)p\left({\bf z}|{\bf i}\right)$          & $q\left({\bf z}|{\bf v}\right)$          & & \Checkmark & 65.6 & 66.1\\
  \bottomrule
\end{tabular}
}
}

%% file: results_external.tex
\scalebox{.8}{
\resizebox{\textwidth}{!}{
\begin{tabular}{lccccc}
  \toprule
  Model & Image res. & Output  & IoU$\uparrow$ & CD$\downarrow$ & EMD$\downarrow$\\
  \midrule
  3D-R2N2~\cite{choy16eccv}      & $127^2$ & voxels $32^3$                          & 56.0       & 7.10$^\dagger$          & 10.20$^\dagger$\\
  OGN~\cite{tatarchenko17iccv}   & $137^2$ & voxel octrees $32^3$                   & 59.6       & ---          & ---\\
  PSGN~\cite{su17cvpr}           & $128^2$ & points $1024$                          & 64.0       & {\bf 2.50}              & 8.00\\
  AtlasNet~\cite{groueix18cvpr}  & $224^2$ & points $2500$                          & ---        & 5.11       & ---\\
  Pixel2Mesh~\cite{wang18eccv}   & $224^2$ & meshes $2466$                          & ---        & 5.91       & 13.80\\
  3D-LMNet~\cite{mandikal18bmvc} & $128^2$ & points $2048$                          & ---        & 5.40       & 7.00\\
  PRN (ours)                     & $137^2$ & voxels $32^3$                          & {\bf 66.2} & 4.42       & {\bf 6.32}\\
  \bottomrule
\end{tabular}
}
}

%% file: conclusion.tex
\section{Conclusion}
\label{sect:conclusion}

In this paper we presented Probabilistic Reconstruction Networks, a generic probabilistic framework for 3D shape inference from single image. 
This framework naturally decouples different aspects of the problem, including the shape  representation, the image conditioning structure, and  the usage of the latent shape space.  
In our experiments with voxel-grid shape representations, we systematically explored the impact of  image conditioning, Monte Carlo \vs variational likelihood approximation for training, the stochastic nature of the latent variable, and joint training with a generative shape model. 
We obtained  single image shape reconstruction results that surpass the previous state of the art in terms of the IoU and EMD performance metrics, and outperform recent work based on point-cloud and mesh-based shape representation.

Given the interpretation of the inference-based reconstruction as an instance of conditional generation, future work includes further adaptation of the generative modelling approaches to the task, as well as the investigation of different shape representations within the proposed framework.

\subsection*{Acknowledgements}
This work has been partially supported by the grant ``Deep in France'' (ANR-16-CE23-0006) and LabEx PERSYVAL-Lab (ANR-11-LABX-0025-01). 

%% file: appendices.tex
\appendix
\section{Histograms of IoU for selected methods}
\label{app:histograms}

\begin{figure}[h!]
    \centering
    \includegraphics[width=.7\textwidth]{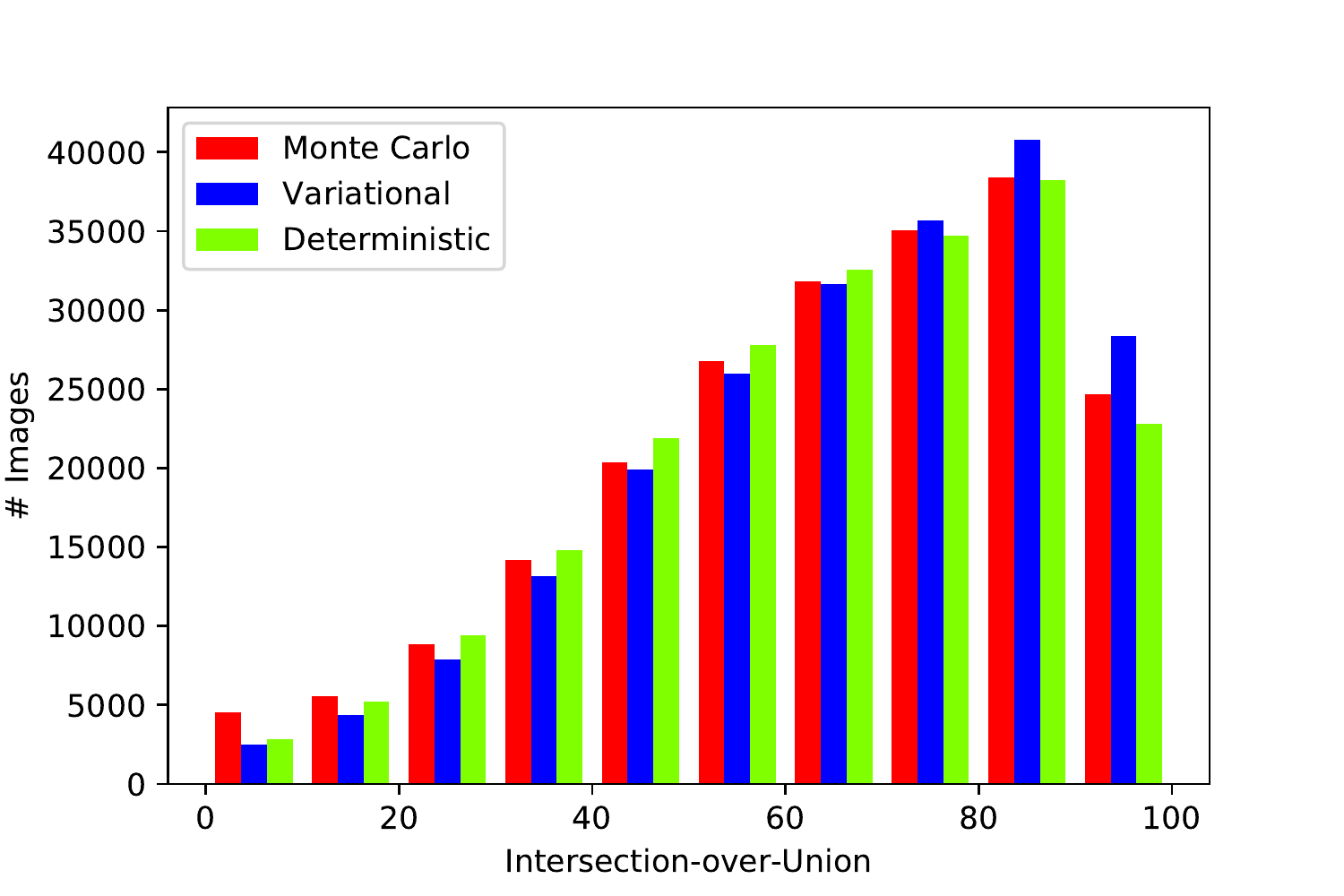}
    \caption{Histogram of IoU values on the ShapeNet test set for the Monte Carlo, variational, and deterministic model. See text for details. Bins of size 10 from 0 to 10, then 10 to 20, \etc}
    \label{fig:hist}
\end{figure}

In \fig{hist} we provide a histogram of the IoU scores obtained across the shapes in the ShapeNet test set using three of the model evaluated in \tab{res_int}:
\begin{itemize}[noitemsep,topsep=0pt]
    \item Using $p({\bf v}|{\bf z})p({\bf z}|{\bf i})$, with Monte Carlo training (\tab{res_int}, row 1).
    \item Using $p({\bf v}|{\bf z})p({\bf z}|{\bf i})$, with variational training (\tab{res_int}, row 4).
    \item Using $p({\bf v}|{\bf z})p({\bf z}|{\bf i})$, with deterministic modeling (\tab{res_int}, row 7).
\end{itemize}
For each shape in the test set there are 24 views, giving a total of about 210k shape inferences.

The histograms show that the variational learning approach leads to more accurate reconstructions, 
leading to the largest number of reconstructed shapes in the last three bins for shape with $>70\%$ IoU.
For all other bins of less accurate results, the variational method has the smallest number of shapes.

Compared to the deterministic model, Monte Carlo training leads to more accurate reconstructions, but also to more very poor reconstructions.  

\section{Visualization of shape reconstruction results}
\label{app:vis}

In this section we provide visualization of additional shape reconstruction results, similar to the ones presented in \sect{exp}. Contrary to them, we put randomly sampled examples here.

\begin{figure}
    \centering
    \scalebox{0.9}{
      \includegraphics[width=\textwidth]{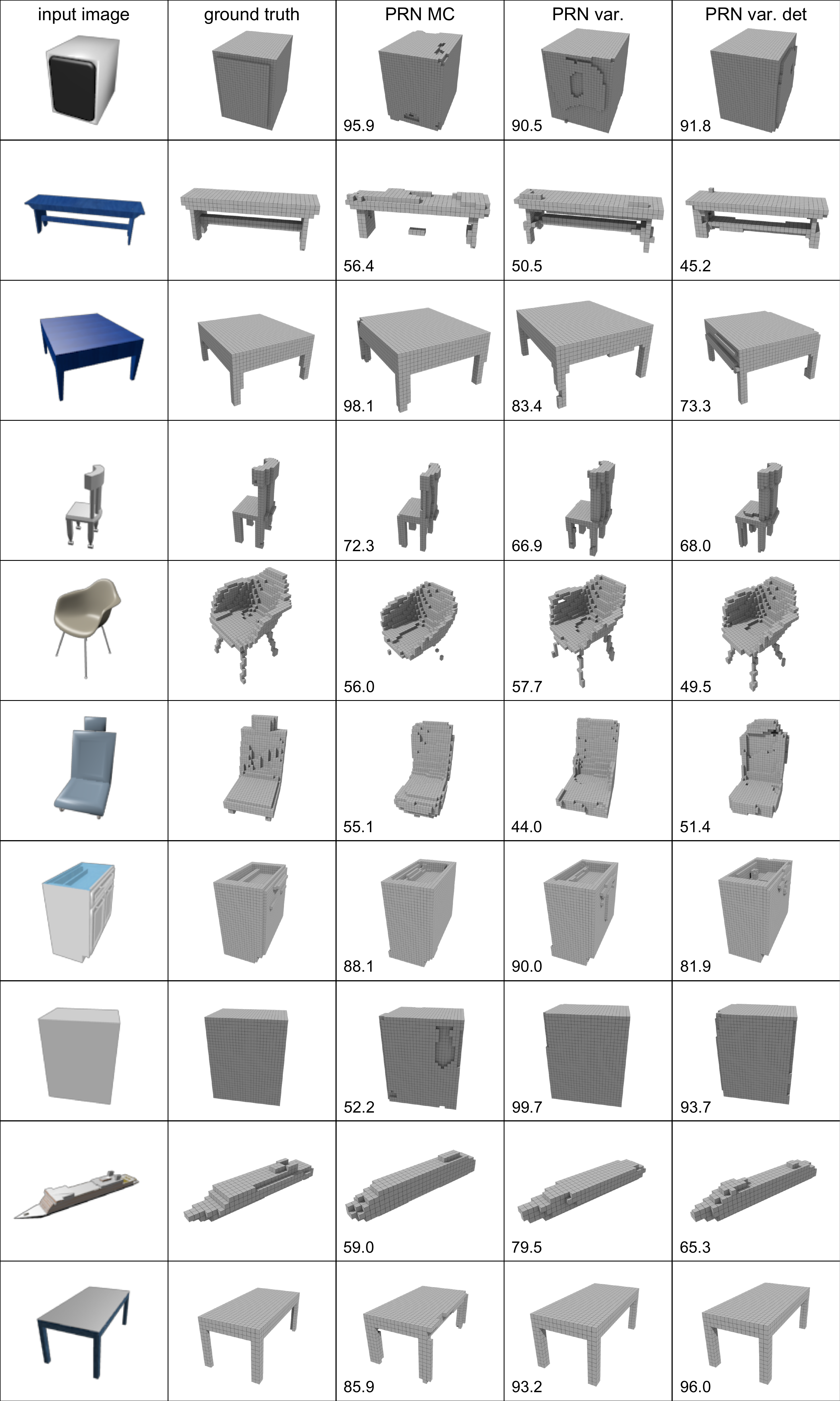}
    }
    \caption{Reconstruction results for random input images from the test set.}
    \label{fig:examples_random}
\end{figure}

For each example in \fig{examples_random}, we show from left to right: 
\begin{itemize}[noitemsep,topsep=0pt]
    \item the input image;
    \item ground-truth shape;
    \item  inferred shape with Monte Carlo training;
    \item inferred shape with variational training;
    \item inferred shape with deterministic model.
\end{itemize}
These shape inference approaches correspond to rows one, four, and seven of \tab{res_int}.